\documentclass{article}
\PassOptionsToPackage{dvipsnames}{xcolor}
\usepackage{iclr2027_conference,times}

\usepackage[dvipsnames]{xcolor}
\definecolor{Gray}{gray}{0.9}
\definecolor{mygreen}{rgb}{0.0, 0.5, 0.0}
\definecolor{myred}{rgb}{0.8, 0.25, 0.33}
\definecolor{myblue}{rgb}{0.19, 0.55, 0.91}
\definecolor{uclablue}{rgb}{0.15, 0.45, 0.68}
\definecolor{ucladblue}{rgb}{0.0, 0.33, 0.53}
\definecolor{ucladdblue}{rgb}{0.0, 0.23, 0.36}
\definecolor{uclagold}{rgb}{1.0, 0.82, 0.0}
\definecolor{ucladgold}{rgb}{1.0, 0.78, 0.17}
\definecolor{ucladdgold}{rgb}{1.0, 0.72, 0.11}
\definecolor{boxgreen}{rgb}{0.02, 0.66, 0.02}
\definecolor{boxred}{rgb}{0.66, 0.1, 0.1}
\definecolor{boxblue}{rgb}{0.01, 0.01, 0.73}

\usepackage{etoc}
\usepackage{lipsum}
\usepackage{wrapfig}
\usepackage{multicol}
\usepackage{mdframed}

\newcommand{\venue}[1]{{\scriptsize\textcolor{gray}{[#1]}}}

\usepackage{mathtools}
\usepackage{amsthm}

\usepackage[utf8]{inputenc}
\usepackage[T1]{fontenc}
\usepackage{url}
\usepackage{amsfonts}
\usepackage{nicefrac}
\usepackage[nopatch=footnote]{microtype}
\usepackage{graphicx}
\usepackage{amsmath,amssymb}
\usepackage{acronym}
\usepackage{enumitem}
\usepackage{balance}
\usepackage{xspace}
\usepackage{setspace}
\usepackage{caption}
\usepackage{booktabs,tabularx,colortbl,multirow,array,makecell}
\usepackage{pifont}
\usepackage{xr-hyper}
\usepackage{tcolorbox}
\tcbuselibrary{breakable}
\usepackage{newtxmath}
\usepackage{dsfont}

\usepackage{silence}
\usepackage{hyperref}
\hypersetup{
    pdftitle={MemVLN: Streaming Episodic Memory for Vision-Language Navigation},
    pdfauthor={},
    breaklinks=true,
    citecolor=ucladblue,
    colorlinks=true,
    linkcolor=ucladblue
}
\usepackage[capitalise]{cleveref}

\usepackage{siunitx}   %
\usepackage{soul}   %
\usepackage{listings}
\definecolor{codegreen}{rgb}{0,0.6,0}
\definecolor{codegray}{rgb}{0.5,0.5,0.5}
\definecolor{codepurple}{rgb}{0.58,0,0.82}
\definecolor{backcolour}{rgb}{0.95,0.95,0.92}

\lstdefinestyle{mystyle}{
    backgroundcolor=\color{backcolour},   
    commentstyle=\color{codegreen},
    keywordstyle=\color{magenta},
    numberstyle=\tiny\color{codegray},
    stringstyle=\color{codepurple},
    basicstyle=\ttfamily\footnotesize,
    breakatwhitespace=false,         
    breaklines=true,                 
    captionpos=b,                    
    keepspaces=true,                 
    numbers=left,                    
    numbersep=5pt,                  
    showspaces=false,                
    showstringspaces=false,
    showtabs=false,                  
    tabsize=2
}

\makeatletter
\DeclareRobustCommand\onedot{\futurelet\@let@token\@onedot}
\def\@onedot{\ifx\@let@token.\else.\null\fi\xspace}
\def\eg{\emph{e.g}\onedot}

\makeatother

\crefname{algorithm}{Alg.}{Algs.}
\Crefname{algocf}{Algorithm}{Algorithms}
\crefname{section}{Sec.}{Secs.}
\Crefname{section}{Section}{Sections}
\crefname{table}{Tab.}{Tabs.}
\Crefname{table}{Table}{Tables}
\crefname{figure}{Fig.}{Fig.}
\Crefname{figure}{Figure}{Figure}

\newcommand{\sota}{state-of-the-art\xspace}
\newcommand{\model}{\textsc{MemVLN}\xspace}
\newcommand{\cmark}{\ding{51}}

\acrodef{llm}[LLM]{large language model}
\acrodef{vla}[VLA]{vision-language-action}
\acrodef{vln}[VLN]{vision-language navigation}
\acrodef{qa}[QA]{question answering}
\acrodef{cot}[CoT]{chain-of-thought}
\acrodef{vlm}[VLM]{vision-language model}
\acrodef{rl}[RL]{reinforcement learning}
\acrodef{sft}[SFT]{supervised fine-tuning}
\acrodef{rlhf}[RLHF]{reinforcement learning from human feedback}
\acrodef{rope}[RoPE]{rotary position embedding}
\acrodef{nlp}[NLP]{natural language processing}
\acrodef{dpo}[DPO]{direct preference optimization}
\acrodef{grpo}[GRPO]{group relative policy optimization}
\acrodef{nll}[NLL]{negative log likelihood}

\title{\model: Streaming Episodic Memory for Vision-Language Navigation}

\iclrfinalcopy

\author{
  Yan Wang$^{1*}$, Jiangyong Huang$^{1,2,*}$, Siyuan Huang$^{1,\dagger}$, Baoxiong Jia$^{1,\dagger}$ \\
  $^{1}$ State Key Laboratory of General Artificial Intelligence, BIGAI \\
  $^{2}$ Peking University \\
  \url{https://memvln.github.io/} \\
  {\footnotesize $^{*}$ Equal contribution. \quad $^\dagger$ Corresponding authors.}
}

\begin{document}
\maketitle


\begin{figure}[htbp]
    \centering
        \includegraphics[width=\linewidth]{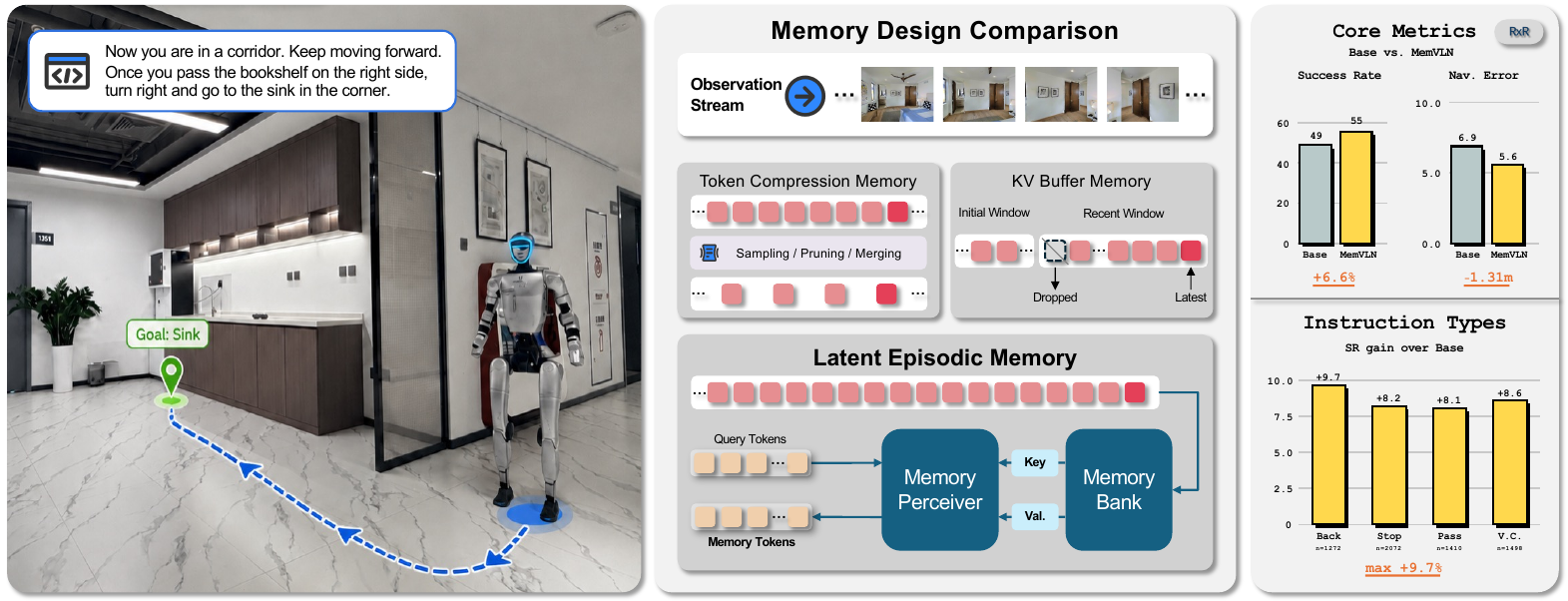}
        \caption{
        \textbf{\model overview.} We design a streaming episodic memory framework that endows \acp{vlm} with navigation memory for streaming video. \model employs a memory module that progressively stores compressed perceptual features and produces compact memory tokens via cross-attention. \model elevates vision-language navigation performance, including long-horizon and real-world scenarios.
        }
    \label{fig:teaser}
    \vspace{-3pt}
\end{figure}

\begin{abstract}
    \Ac{vln} requires an embodied agent to follow language instructions through 3D environments. Maintaining persistent episodic memory remains a challenge for \ac{vlm}-based streaming agents in long-horizon navigation. We present \model, a streaming episodic memory framework that maintains a bounded bank of compressed perceptual features annotated with timesteps and action-derived relative poses, and produces compact memory tokens via cross-attention. We pair the proposed memory with a frozen pretrained geometry encoder. To learn the memory module, we design a masking-based trajectory summarization task with multi-granularity sub-trajectory decomposition, encouraging route reconstruction exclusively from episodic memory. Under navigation training restricted to the official R2R/RxR trajectories, \model achieves \sota performance on most reported metrics on standard VLN-CE benchmarks. Long-horizon analyses further characterize when episodic memory is useful. For physical evaluation, we collect 30 instruction-conditioned episodes in one physical scene, use 24 as demonstrations for post-training, and hold out six for testing. On the held-out episodes, \model improves success over the \ac{vlm} baseline from 17\% to 40\%, and post-training further raises it to 47\%.
\end{abstract}


\section{Introduction}
\label{sec:introduction}

\Acf{vln} is a fundamental task in embodied AI, requiring an embodied agent to follow language instructions through 3D environments~\citep{anderson2018vision,krantz2020beyond,majumdar2023we}. Recent \acfp{vlm} have substantially improved navigation performance, with monocular streaming video as a prevailing formulation that is natural and practical for real-world deployment~\citep{zhang2025uninavid,cheng2025navila,wei2026streamvln,wei2026ground}. Despite rapid progress, the lack of persistent memory remains a critical constraint for real-world long-horizon navigation, where decisions depend on observations from earlier in the episode. In this paper, we address the key problem: \textit{how to design a simple yet effective episodic memory that integrates seamlessly with streaming \ac{vlm} to elevate \ac{vln} capability}?  

Navigation memory has received prior attention, yet existing approaches remain insufficient along several key dimensions. Early methods encode navigation history in recurrent hidden states~\citep{elman1990finding,hochreiter1997long,hong2021vln}, compressing perceptual observations into a fixed-size representation. Explicit memory representations retain observation-level features but are often coupled to panoramic observations or topological maps, requiring adaptation to monocular streaming video in continuous environments~\citep{chen2021history,mezghan2022memory,an2024etpnav,wang2023gridmm,yu2025mossvln}. More recent works revisit navigation memory within a streaming formulation~\citep{zeng2026janusvln,zheng2025efficient}, yet the capacity limitation persists. These shortcomings motivate our approach: \textit{an episodic memory design that is expressive and natively compatible with streaming video, accompanied by a systematic study of navigation memory spanning long-horizon navigation scenarios and real-world validation}.

We propose \model, a streaming episodic memory framework that endows \acp{vlm} with persistent navigation memory. \model maintains dense short-term observations for immediate action while building a compact, queryable summary of the accumulated episode history. Concretely, a memory bank progressively stores perceptual features of streaming observations alongside their timestep and action-derived relative pose. For action prediction, a memory perceiver extracts latent features from the bank and encodes them with the corresponding spatio-temporal context. It then derives a fixed-size set of memory tokens via cross-attention with learnable queries grounded in the current timestep and relative pose. Memory tokens enter the \ac{vlm}'s multi-modal sequence for auto-regressive action prediction. When the bank reaches its capacity limit, a frame-merging strategy consolidates consecutive frames into a bounded episodic summary. As a complementary source of spatial features, we use a frozen pretrained StreamVGGT encoder~\citep{zhuo2026streaming} and train a lightweight adapter to fuse its features into the \ac{vlm}. Both the memory module and the geometry encoder integrate into existing \acp{vlm} for streaming video without modifying the underlying LLM architecture.

To supervise memory learning, we design a dedicated auxiliary task alongside navigation. Specifically, we adopt trajectory summarization~\citep{zheng2024towards,zhang2024navid,cheng2025navila} as the supervision for the memory module, where local visual tokens are masked, leaving memory tokens as the visual evidence for describing the traversed path. To enrich the supervisory signal, we further decompose each full trajectory into hierarchically structured sub-trajectories~\citep{hong2020sub}, yielding fine-grained supervision across multiple temporal scales. Together, the memory module design, multi-granularity trajectory summarization, and masking strategy constitute a principled framework for learning navigation memory.

We conduct extensive experiments in simulation and on a physical robot. On standard VLN-CE benchmarks~\citep{anderson2018vision,ku2020room,krantz2020beyond}, \model reaches at least 55\% success on both R2R and RxR and achieves \sota performance on most metrics among methods whose navigation fine-tuning trajectories are restricted to the official R2R/RxR training splits. Ablations isolate the contributions of memory, geometry, and auxiliary supervision, while long-horizon analyses further examine when memory is useful. For physical evaluation on the G1 robot, we collect 30 episodes in one physical scene, use 24 as demonstrations for post-training, and hold out six for testing. On the held-out episodes, \model improves success over the \ac{vlm} baseline from 17\% to 40\%, and post-training further raises it to 47\%.


\section{Related Work}
\label{sec:related_work}

\paragraph{Vision-Language Navigation.}
\Ac{vln} requires an embodied agent to follow natural language instructions through 3D environments. Early works establish \ac{vln} as a foundational task for embodied AI~\citep{chang2017matterport3d,zhu2017target,xia2018gibson,anderson2018evaluation,savva2019habitat,anderson2018vision,chaplot2020object,qi2020reverie,ku2020room,krantz2020beyond,thomason2020vision,ramakrishnan2021habitat,krantz2023navigating,yadav2023habitat}. Given scarce annotated navigation data, considerable work targets data-centric approaches, including data augmentation~\citep{tan2019learning,liu2021vision,zhu2020babywalk,hong2020sub,he2025multilevel,cui2025generating} and data scaling in both simulation environments~\citep{hao2020towards,kamath2023new,wang2023scaling} and the real world~\citep{guhur2021airbert,linghu2024multi,cheng2025navila,chen2026lifting}.
As model capabilities advance, \ac{vln} evaluation encompasses increasingly complex scenarios~\citep{majumdar2024openeqa,zhang2024task,yokoyama2024hm3d,khanna2024goat,qiao2025navbench,lin2025vlnverse,wang2025rethinking}. In this paper, we focus on the fundamental \ac{vln} task in continuous environments and study how navigation memory can be learned and leveraged without relying on massive data scaling~\citep{anderson2018vision,ku2020room,krantz2020beyond,hong2020sub}.

\paragraph{Vision-Language Models for Navigation.}
Effective \ac{vln} requires models to perceive 3D environments and comprehend natural language instructions. While early approaches trained navigation models from scratch~\citep{fried2018speaker,nguyen2019vision,wang2019reinforced,hao2020towards}, the strengths of pretrained \acp{vlm} have emerged as a more promising pathway towards general-purpose \ac{vln}~\citep{majumdar2022zson,gadre2023cows,dorbala2022clip,zhou2023esc,wang2024navigating,long2024instructnav,zhou2024navgpt}. This paradigm has inspired diverse approaches, including topological structure methods~\citep{chen2021topological,krantz2021waypoint,hong2022bridging,chen2022weakly,chen2022think,yokoyama2024vlfm,an2024etpnav}, explicit 3D spatial modeling~\citep{an2023bevbert,wang2024sim,huang2024embodied,zhu2025move,wang2025dynam3d,wang2026d3d}, and trajectory-level modeling~\citep{shah2022viking,shah2023vint,sridhar2024nomad,wang2025trackvla,gode2025flownav}. More recently, the input formulation has been simplified to monocular streaming video, under which navigation performance has improved rapidly~\citep{zheng2024towards,zhang2024navid,zhang2025uninavid,cheng2025navila,wei2026streamvln,zeng2026janusvln,zheng2025efficient,qi2025vln,wei2026ground,chu2026abot}, with significant progress also demonstrated in real-world deployment~\citep{gervet2023navigating,yue2024safe,yang2024pushing,qiao2025open,sun2026view,yue2026spatial,yu2026correctnav,zhang2026embodied}. Despite this progress, directly retaining long visual histories in streaming \ac{vlm} policies increases input context and computation, motivating dedicated memory mechanisms.

\paragraph{Memory for Embodied Agents.}
Memory mechanisms for embodied agents predate modern \ac{vlm}-based navigation and span several task formulations. Recurrent policies compress navigation history into hidden states~\citep{elman1990finding,hochreiter1997long,hong2021vln}, whereas explicit approaches retain observation histories or structured spatial representations~\citep{chen2021history,mezghan2022memory,wang2023gridmm,yu2025mossvln}. SMT appends each observation to a growing within-episode memory for visual navigation, while E.T. jointly encodes language with the full visual-action history for ALFRED~\citep{fang2019scene,pashevich2021episodic}. ESceme instead maintains a cross-episode, location-indexed scene memory for VLN~\citep{zheng2025esceme}. Beyond navigation, MemoryVLA retrieves decision-relevant history from a perceptual-cognitive memory for robotic manipulation~\citep{shi2026memoryvla}. In contrast, MemVLN targets a bounded, episode-local visual bank for streaming VLN, with time/pose-grounded latent readout and memory-specific trajectory-summary supervision.


\section{Method}
\label{sec:method}
\paragraph{Task Formulation.} We study vision-language navigation in continuous environments (VLN-CE), where an embodied agent receives streaming observations and follows language instructions through 3D environments. At each timestep $t$, the agent receives an egocentric observation $o_t$ and a task instruction $\mathcal{T}$, and predicts an action $a_t$. The objective is to learn a navigation policy $p(a_t \mid o_{1:t}, \mathcal{T})$. Following prior work~\citep{cheng2025navila}, actions are represented as text comprising an action type $a_t^\text{type} \in$ \{\texttt{move~forward}, \texttt{turn~left}, \texttt{turn~right}, \texttt{stop}\} and an action scale $a_t^\text{scale}$, where forward distances are in \{25, 50, 75\}cm and turn angles are in \{$\ang{15}$, $\ang{30}$, $\ang{45}$\}.

\subsection{Model Overview}
\label{sec:model_overview}

\begin{figure}[t!]
    \centering
    \includegraphics[width=\linewidth]{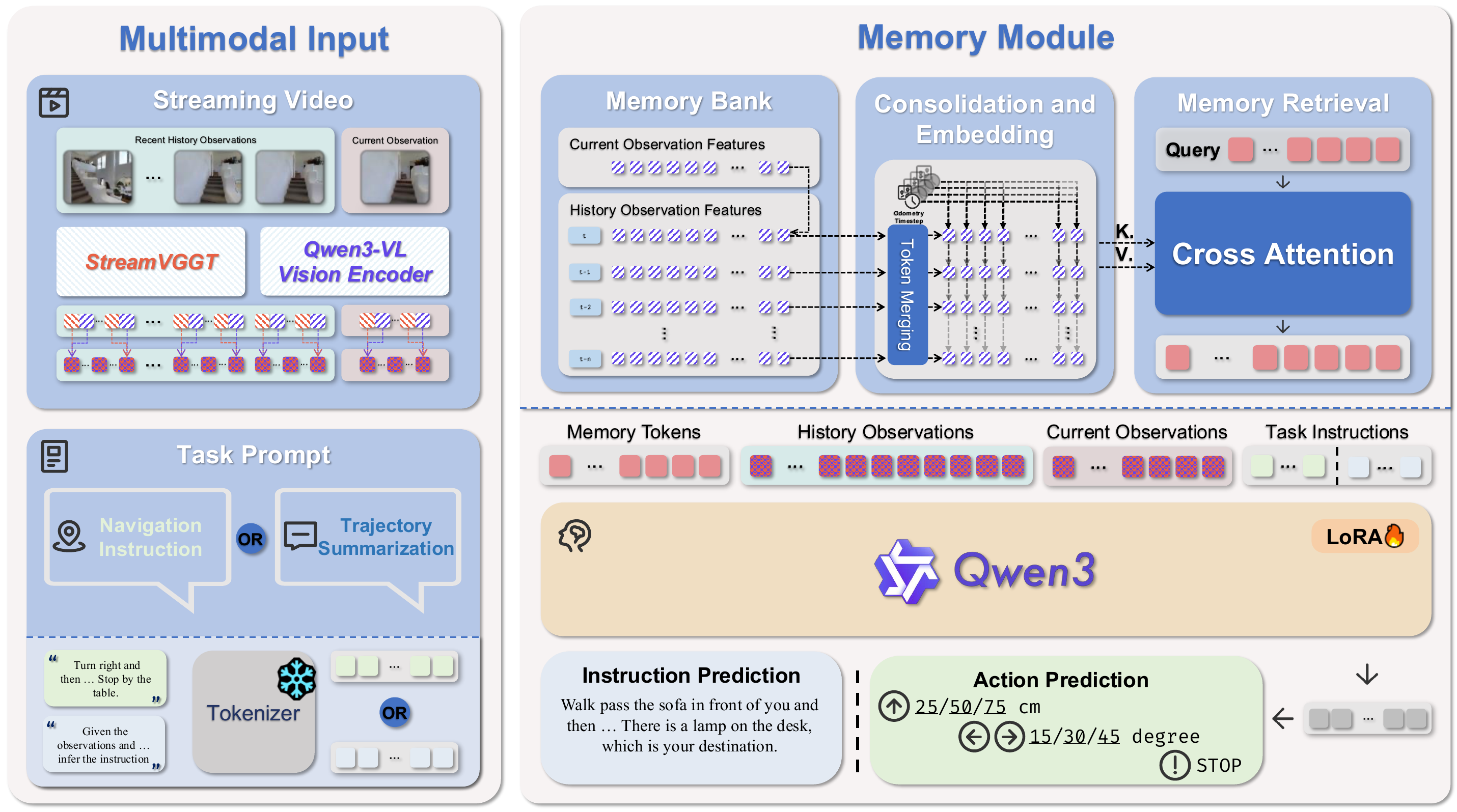}
    \caption{\textbf{\model model overview.} \model combines Qwen3-VL's native SigLIP-2-based vision encoder and LLM, a frozen StreamVGGT geometry encoder, and our memory module. It receives a short-term context for local perception and an episodic memory for long-horizon reasoning.}
    \label{fig:method}
    \vspace{-3pt}
\end{figure}

As illustrated in~\cref{fig:method}, \model comprises the native vision encoder and \ac{llm} of Qwen3-VL, a frozen geometry encoder, and a memory module. At navigation step $t$, the agent acquires a new RGB observation $o_t\in\mathds{R}^{H\times W\times 3}$, which together with recent observations forms the short-term visual context. The native SigLIP-2-based vision encoder $\pi_V$ encodes this context into visual features, with $f_t\in\mathds{R}^{H'\times W'\times d}$ denoting the current observation's patch features. We use pretrained visual geometry models to improve the base \ac{vlm}'s spatial understanding. In parallel, frozen StreamVGGT~\citep{wang2025vggt,zhuo2026streaming} provides camera-token representations for the same context. A lightweight trainable adapter $\phi$ aligns and fuses the geometry and visual representations, producing local hybrid tokens $\mathcal{H}_t=[h_t^{\text{hist}},h_t]$ for the recent history and current observation, respectively. Complementing the short-term visual context, the memory module encodes the accumulated visual history into memory tokens $m_t$, as detailed in~\cref{sec:memory}.


\paragraph{Vision-Language Model Design.} \model augments the vanilla \ac{llm} input with memory tokens and geometry features, modeling $p(a_t \mid m_t, \mathcal{H}_t, \mathcal{T})$. To enhance awareness of the current situation, we explicitly distinguish the recent-history tokens $h_t^{\text{hist}}$ from the current-observation tokens $h_t$ within $\mathcal{H}_t$. The interleaved multi-modal input to the \ac{llm} $\pi_L$ is:
\begin{equation}
\begin{split}
\Big[
\big\{\underbrace{\mathcal{P}_{\text{mem}}}_{\substack{\text{memory} \\ \text{prompt}}}, \, 
\underbrace{m_t}_{\substack{\text{memory} \\ \text{tokens}}}\big\}, \, 
\big\{\underbrace{\mathcal{P}_{\text{hist}}}_{\substack{\text{history} \\ \text{observation} \\ \text{prompt}}}, \, 
\underbrace{h_t^{\text{hist}}}_{\substack{\text{history} \\ \text{observation} \\ \text{tokens}}}\big\}, \,
\big\{\underbrace{\mathcal{P}_{\text{curr}}}_{\substack{\text{current} \\ \text{observation} \\ \text{prompt}}}, \, 
\underbrace{h_t}_{\substack{\text{current} \\ \text{observation} \\ \text{tokens}}}\big\}, \, 
\underbrace{\mathcal{T}}_{\substack{\text{task} \\ \text{instruction}}}, \, 
\underbrace{a_t}_{\substack{\text{action} \\ \text{output}}} \Big],
\end{split}
\end{equation}
\begin{equation}
    a_t = \pi_L(m_t,\mathcal{H}_t,\mathcal{T}).
\end{equation}
The history observation components $\mathcal{P}_{\text{hist}}$ and $h_t^{\text{hist}}$ are omitted at the initial step $t=1$. The dense short-term context anchors fine-grained local perception, while the retrieved memory supplies a bounded summary of accumulated navigation history.

\subsection{Memory Module}\label{sec:memory}

\paragraph{Memory Bank.} Upon receiving a new observation $o_t$, the corresponding visual feature $f_t$ is appended to an episode-specific memory bank along with its timestep $t$ and relative pose $z_t$. The pose encodes position $(x,y)$ and orientation $(\cos\theta,\sin\theta)$ relative to the starting point. At policy inference, it is dead-reckoned solely from the executed discrete-action history; neither ground-truth pose nor external odometry is provided to the policy or memory module. When the bank reaches its maximum capacity, the two consecutive frames with the highest average patch-level cosine similarity are merged, retaining a bounded representative summary. We store the visual feature $f_t$ rather than the hybrid feature $h_t$.

\paragraph{Memory Perceiver.} A memory perceiver resamples the bank into a compact representation using learnable queries and cross-attention~\citep{vaswani2017attention,jaegle2021perceiver}. Prior to serving as keys and values, the bank features undergo two operations:
\begin{itemize}
    \item \textbf{Spatial Downsampling.} To reduce computational cost, patch features within each frame are spatially merged using ToMe~\citep{bolya2023token}:
    \begin{equation}
        \bar{f}_{1:t} = \texttt{ToMe}(f_{1:t}).
    \end{equation}
    \item \textbf{Spatio-temporal Embedding.} Timestep and action-derived relative pose are encoded into both the bank features and learnable queries $q$ via sinusoidal temporal embeddings~\citep{vaswani2017attention} and Fourier spatial features~\citep{rahimi2007random,tancik2020fourier,li2021learnable}:
    \begin{equation}
        \bar{f}_{1:t}^{+} = \bar{f}_{1:t} + \texttt{sinusoid}(1:t) + \texttt{Fourier}(z_{1:t}),
    \end{equation}
     \begin{equation}
        q^+ = q + \texttt{sinusoid}(t) + \texttt{Fourier}(z_{t}).
    \end{equation}
\end{itemize}
The memory perceiver then yields a compact and fixed-size set of memory tokens via cross-attention:
\begin{equation}
    m_t = \texttt{cross-attn}(q^+, \bar{f}_{1:t}^{+}).
\end{equation}
The memory module offers three complementary properties for \ac{vln}: (1) \textbf{Long-horizon context}: the bank retains representative visual evidence (\eg, signs, doorways, landmarks) from across the episode, extending the available context beyond recent observations; (2) \textbf{Bounded representation}: frame consolidation bounds the number of stored entries, spatial downsampling limits the tokens processed per frame, and the perceiver produces a fixed-size set of memory tokens; and (3) \textbf{Spatio-temporal grounding}: timestep and action-derived relative-pose embeddings relate stored evidence to the agent's current point along the trajectory. At inference, the readout queries are instruction-agnostic by design: the memory serves as an episodic record of what the agent has experienced, while the downstream \ac{llm} performs instruction-conditioned interpretation.

\subsection{Data and Training}
\label{sec:data_training}

\model is trained jointly on two tasks: \ac{vln} and trajectory summarization. The navigation task teaches the model to predict the next action, while the summarization task shapes the memory representation towards route-level semantics. Both tasks use a unified cross-entropy loss.

\paragraph{Vision-Language Navigation.} \Ac{vln} trajectories are preprocessed with action merging following NaVILA~\citep{cheng2025navila}, where consecutive atomic actions within a window of three steps are merged, \eg, three consecutive ``\texttt{move forward 25cm}'' actions become a single ``\texttt{move forward 75cm}''. This enriches action diversity and improves training efficiency. At each step, the model is supervised to predict the ground-truth action $a_t^*$:
\begin{equation}
    \mathcal{L} = - \log p(a_t^* \mid m_t,\mathcal{H}_t,\mathcal{T}).
\end{equation}

\paragraph{Trajectory Summarization.} Trajectory summarization provides auxiliary supervision for the memory module. Using the same \ac{vln} trajectories, the task instruction $\mathcal{T}$ is repurposed as the ground-truth label. During summarization, all local visual tokens $\mathcal{H}_t$ are masked, leaving the memory tokens $m_t$ as visual evidence. This masking strategy trains the model to reconstruct navigational semantics exclusively from episodic memory, supervising it to retain evidence relevant to long-horizon instruction following, such as landmarks marking completed subgoals or visual cues indicating stopping positions. With the summarization prompt $\mathcal{T_\text{(TS)}}$, the loss is:
\begin{equation}
    \mathcal{L} = - \log p(\mathcal{T} \mid m_t,\mathcal{T_\text{(TS)}}).
\end{equation}

\paragraph{Sub-trajectory Derivation.} We enrich whole-trajectory summarization with sub-trajectory supervision. FGR2R annotations~\citep{hong2020sub} provide sub-instructions and corresponding waypoints in discrete navigation graphs. We map these waypoints to continuous environments by retrieving the nearest points on the continuous trajectory. This yields sub-trajectory splits that provide finer-grained supervision, encouraging the memory module to capture semantics at multiple temporal scales.


\section{Experiments}
\label{sec:experiment}

\begin{table}[t!]
\centering
\caption{
\textbf{Evaluation results on VLN-CE R2R Val-Unseen split.} ``Pano.'', ``Odo.'', and ``S.RGB'' denote panoramic view, odometry, and single RGB; at policy inference, MemVLN instead derives relative pose from executed actions. External Data counts fine-tuning examples beyond those derived from official R2R/RxR trajectories (\eg, EnvDrop~\citep{tan2019learning}). Bold marks the best streaming result among methods restricted to those official navigation trajectories. Results are partially referenced from JanusVLN~\citep{zeng2026janusvln}.
}
\label{tab:main_results}
\resizebox{\textwidth}{!}{
\begin{tabular}{l|cccc|cccc|r}
\toprule
\multirow{2}{*}[-0.3em]{Method}
&
\multicolumn{4}{c|}{Observation}
&
\multicolumn{4}{c|}{R2R Val-Unseen}
&
\multicolumn{1}{c}{Training}
\\
\cmidrule(lr){2-5}
\cmidrule(lr){6-9}
\cmidrule(lr){10-10}
&
Pano. & Odo. & Depth & S.RGB
&
NE$\downarrow$ & OS$\uparrow$ & SR$\uparrow$ & SPL$\uparrow$
&
External Data
\\
\midrule

CMA~\citep{hong2022bridging} \venue{CVPR22}
& \cmark & \cmark & \cmark &  & 6.20 & 52.0 & 41.0 & 36.0 & - \\

VLN$\circlearrowright$BERT~\citep{hong2021vln} \venue{CVPR21} 
& \cmark & \cmark & \cmark &  & 5.74 & 53.0 & 44.0 & 39.0 & - \\

Ego$^2$-Map~\citep{hong2023learning} \venue{ICCV23} 
& \cmark & \cmark & \cmark &  & 5.54 & 56.0 & 47.0 & 41.0 & - \\

DreamWalker~\citep{wang2023dreamwalker} \venue{ICCV23} 
& \cmark & \cmark & \cmark &  & 5.53 & 59.0 & 49.0 & 44.0 & - \\

\midrule

AO-Planner~\citep{chen2025affordances} \venue{AAAI25} 
& \cmark &  & \cmark &  & 5.55 & 59.0 & 47.0 & 33.0 & - \\

g3D-LF~\citep{wang2025g3d} \venue{CVPR25} 
&  & \cmark & \cmark & \cmark & 5.70 & 59.5 & 47.2 & 34.6 & - \\


NaVid-4D~\citep{liu2025navid} \venue{ICRA25} 
&  &  & \cmark & \cmark & 5.99 & 55.7 & 43.8 & 37.1 & - \\

NavMorph~\citep{yao2025navmorph} \venue{ICCV25} 
&  &  & \cmark & \cmark & 5.75 & 56.9 & 47.9 & 33.2 & - \\

\midrule

NaVid~\citep{zhang2024navid} \venue{RSS24} 
&  &  &  & \cmark & 5.47 & 49.1 & 37.4 & 35.9 & 953$K$ \\

Sim2Real~\citep{wang2024sim} \venue{CoRL24} 
&  &  &  & \cmark & 5.95 & 55.8 & 44.9 & 30.4 & 0$K$ \\

StreamVLN~\citep{wei2026streamvln} \venue{ICRA26}
&  &  &  & \cmark & 6.05 & 53.8 & 45.5 & 41.6 & 10033$K$ \\

Uni-NaVid~\citep{zhang2025uninavid} \venue{RSS25} 
&  &  &  & \cmark & 5.58 & 53.3 & 47.0 & 42.7 & 3577$K$ \\

NaVILA~\citep{cheng2025navila} \venue{RSS25}
&  &  &  & \cmark & 5.37 & 57.6 & 49.7 & 45.5 & 12574$K$ \\

JanusVLN~\citep{zeng2026janusvln} \venue{ICLR26}
&  &  &  & \cmark & \textbf{5.17} & 58.0 & 52.8 & 49.2 & 0$K$ \\

\rowcolor{gray!25}
\textbf{MemVLN (Ours)}
&  &  &  &  \cmark & 5.19 & \textbf{61.2} & \textbf{55.0} & \textbf{50.5} & 0$K$ \\




\bottomrule
\end{tabular}
}
\end{table}

\paragraph{Implementation Details.} We use Qwen3-VL-4B~\citep{bai2025qwen3}, including its native SigLIP-2-based vision encoder, together with StreamVGGT-1B~\citep{zhuo2026streaming}. Both encoders are frozen. We adapt the \ac{llm} with LoRA~\citep{hu2022lora} and train the memory module and geometry adapter from scratch, yielding 1B trainable parameters in total. The recent observation window contains up to 16 frames. The model is trained for 1 epoch (2 days on 8$\times$A100 80G GPUs) using AdamW~\citep{loshchilov2019decoupled} with a base learning rate of $2\times 10^{-5}$, linear warmup, and cosine decay. See detailed hyperparameters in \textit{Appendix}.
\paragraph{Data and Evaluation.} We train on the official R2R~\citep{anderson2018vision} and RxR~\citep{ku2020room} training trajectories, comprising 10,819 and 19,996 demonstrations. Full-trajectory summary targets reuse the instructions from these routes; fine-grained R2R supervision uses FGR2R sub-instruction/waypoint annotations derived from R2R routes~\citep{hong2020sub}. No additional navigation trajectories or external multimodal fine-tuning data are used. For simulation, we roll out the policy in Habitat-Sim~\citep{savva2019habitat} under the VLN-CE protocol~\citep{krantz2020beyond} and report Val-Unseen results. For R2R, we report success rate (SR), oracle success rate (OSR), success weighted by path length (SPL), and navigation error (NE). For RxR, we report SR, SPL, normalized dynamic time warping (nDTW)~\citep{ilharco2019general}, and NE.

\subsection{Main Simulation Results}
\label{sec:main_results}

\paragraph{Comparison Baselines.} We compare primarily against recent VLM-based VLN methods, emphasizing the streaming-video formulation~\citep{wei2026streamvln,zeng2026janusvln}. For methods with multiple training configurations, \cref{tab:main_results,tab:rxr_val_unseen} report the selected configuration and its additional training-example count. Our \sota comparison is restricted to methods whose navigation fine-tuning trajectories come only from the official R2R/RxR training splits.


\paragraph{Results.} Under this official-trajectory regime, \model achieves \sota performance on most metrics in \cref{tab:main_results,tab:rxr_val_unseen}. On R2R, \model reaches 55.0\% SR and 50.5\% SPL, exceeding JanusVLN by 2.2 and 1.3 percentage points (pp), respectively. OSR rises from 58.0\% to 61.2\%, with comparable NE (5.19 vs.\ 5.17\,m). On RxR, SR rises from 51.4\% to 55.4\% (+4.0\,pp) and SPL from 44.3\% to 48.4\% (+4.1\,pp). nDTW increases from 59.1 to 64.6, while NE falls from 6.46 to 5.59\,m ($-0.87$\,m). These larger RxR gains highlight \model's effectiveness on longer instructions and more complex trajectories.

\begin{table}[t!]
\centering
\caption{
\textbf{Evaluation results on VLN-CE RxR Val-Unseen split.} Settings follow \cref{tab:main_results} and JanusVLN~\citep{zeng2026janusvln}.
}
\label{tab:rxr_val_unseen}
\small
\setlength{\tabcolsep}{4.0pt}
\renewcommand{\arraystretch}{0.95}

\resizebox{\textwidth}{!}{
\begin{tabular}{l|cccc|cccc|r}
\toprule
\multirow{2}{*}[-0.3em]{Method}
&
\multicolumn{4}{c|}{Observation}
&
\multicolumn{4}{c|}{RxR Val-Unseen}
&
\multicolumn{1}{c}{Training}
\\
\cmidrule(lr){2-5}
\cmidrule(lr){6-9}
\cmidrule(lr){10-10}
&
Pano. & Odo. & Depth & S.RGB
&
NE$\downarrow$ & SR$\uparrow$ & SPL$\uparrow$ & nDTW$\uparrow$
&
External Data
\\
\midrule

CMA~\citep{hong2022bridging} \venue{CVPR22}
& \cmark & \cmark & \cmark &  & 8.76 & 26.5 & 22.1 & 47.0 & - \\

VLN$\circlearrowright$BERT~\citep{hong2021vln} \venue{CVPR21} 
& \cmark & \cmark & \cmark &  & 8.98 & 27.0 & 22.6 & 46.7 & - \\

\midrule

AO-Planner~\citep{chen2025affordances} \venue{AAAI25}
& \cmark &  & \cmark &  & 7.06 & 43.3 & 30.5 & 50.1 & - \\


NavMorph~\citep{yao2025navmorph} \venue{ICCV25}
&  &  & \cmark & \cmark & 8.85 & 30.8 & 22.8 & 44.2 & - \\

\midrule

Sim2Real~\citep{wang2024sim} \venue{CoRL24}
&  &  &  & \cmark & 8.79 & 36.7 & 25.5 & 18.1 & 0$K$ \\

Uni-NaVid~\citep{zhang2025uninavid} \venue{RSS25}
&  &  &  & \cmark & 6.24 & 48.7 & 40.9 & - & 3577$K$ \\

NaViLA~\citep{cheng2025navila} \venue{RSS25}
&  &  &  & \cmark & 6.77 & 49.3 & 44.0 & 58.8 & 13132$K$ \\

JanusVLN~\citep{zeng2026janusvln} \venue{ICLR26}
&  &  &  & \cmark & 6.46 & 51.4 & 44.3 & 59.1 & 0$K$ \\

StreamVLN~\citep{wei2026streamvln} \venue{ICRA26}
&  &  &  & \cmark & 6.22 & 52.9 & 46.0 & 61.9 & $\sim$26330$K$ \\


\rowcolor{gray!25}
\textbf{MemVLN (Ours)}
&  &  &  & \cmark
& \textbf{5.59} & \textbf{55.4} & \textbf{48.4} & \textbf{64.6} & 0$K$ \\

\bottomrule
\end{tabular}
}
\end{table}

\subsection{Ablation Studies}
\label{sec:ablations}

\paragraph{Component Analysis.} We ablate memory (mem.), geometry (geo.), and trajectory-summary supervision (aux.) on R2R-CE and RxR-CE in \cref{tab:ablations}. Memory alone improves success on both benchmarks, raising R2R OSR from 51.1\% to 56.7\% and RxR SR from 48.8\% to 53.3\%. Geometry provides complementary spatial information, increasing SR from 45.6\% to 49.7\% on R2R and from 53.3\% to 56.1\% on RxR when added to memory. The full model exceeds geometry-only by 5.8\,pp SR and 5.3\,pp SPL on R2R, while lowering NE from 6.04 to 5.19\,m. This highlights the value of learning episodic representations alongside spatial perception.

Trajectory-summary supervision adds 5.3\,pp SR and 4.7\,pp SPL over memory plus geometry on R2R. On RxR, the full model achieves the lowest NE and highest nDTW among these variants. These R2R gains suggest that summarizing past trajectories helps memory capture navigation behavior useful for subsequent decisions.

\begin{table}[t!]
    \centering
    \caption{\textbf{Ablations of memory (mem.), geometry (geo.), and auxiliary supervision (aux.).}}
    \label{tab:ablations}
    \setlength{\tabcolsep}{12pt}
    \resizebox{\linewidth}{!}{
    \begin{tabular}{lcccccccc}
        \toprule
        \multirow{2}{*}[-0.3em]{Setting} & \multicolumn{4}{c}{R2R Val-Unseen} & \multicolumn{4}{c}{RxR Val-Unseen} \\
        \cmidrule(lr){2-5} \cmidrule(lr){6-9}
         & NE$\downarrow$ & OS$\uparrow$ & SR$\uparrow$ & SPL$\uparrow$ & NE$\downarrow$ & SR$\uparrow$ & SPL$\uparrow$ & nDTW$\uparrow$ \\
        \midrule
        Baseline & 6.56 & 51.1 & 43.9 & 40.9 & 6.90 & 48.8 & 43.9 & 59.7 \\
        + mem. & 6.38 & 56.7 & 45.6 & 41.6 & 6.28 & 53.3 & 45.9 & 59.6 \\
        + geo. & 6.04 & 58.1 & 49.2 & 45.2 & 5.87 & \textbf{56.3} & \textbf{49.7} & 62.6 \\
        + mem. + geo. & 5.85 & 56.3 & 49.7 & 45.8 & 5.74 & 56.1 & 49.4 & 63.0 \\
        + mem. + geo. + aux. & \textbf{5.19} & \textbf{61.2} & \textbf{55.0} & \textbf{50.5} & \textbf{5.59} & 55.4 & 48.4 & \textbf{64.6} \\
        \bottomrule
    \end{tabular}}
\end{table}

\paragraph{Memory Capacity.} In \cref{tab:memory_capacity}, our default setting ($N_b=100$, $N_m=300$) improves R2R-CE SR by 4.1\,pp and SPL by 3.3\,pp over the zero-memory setting, while achieving the best SR and SPL in both capacity ablations. Results on RxR-CE and configuration details are provided in \cref{sec:supp_memory_capacity}.

\begin{table}[t!]
    \centering
    \caption{\textbf{Memory capacity ablations on R2R Val-Unseen: bank size $N_b$ and token count $N_m$.}}
    \label{tab:memory_capacity}
    \small
    \setlength{\tabcolsep}{3pt}
    \begin{minipage}[t]{0.48\linewidth}
        \centering
        \textbf{(a) Memory bank capacity}\par
        \begin{tabular*}{\linewidth}{@{\extracolsep{\fill}}ccccc@{}}
            \toprule
            $N_b$ & NE$\downarrow$ & OSR$\uparrow$ & SR$\uparrow$ & SPL$\uparrow$ \\
            \midrule
            0 & 5.87 & 57.4 & 50.9 & 47.2 \\
            100 & \textbf{5.19} & \textbf{61.2} & \textbf{55.0} & \textbf{50.5} \\
            200 & 6.24 & 58.7 & 50.2 & 45.7 \\
            \bottomrule
        \end{tabular*}
    \end{minipage}\hfill
    \begin{minipage}[t]{0.48\linewidth}
        \centering
        \textbf{(b) Memory-token count}\par
        \begin{tabular*}{\linewidth}{@{\extracolsep{\fill}}ccccc@{}}
            \toprule
            $N_m$ & NE$\downarrow$ & OSR$\uparrow$ & SR$\uparrow$ & SPL$\uparrow$ \\
            \midrule
            0 & 5.87 & 57.4 & 50.9 & 47.2 \\
            300 & \textbf{5.19} & \textbf{61.2} & \textbf{55.0} & \textbf{50.5} \\
            600 & 6.75 & 53.9 & 45.1 & 40.8 \\
            \bottomrule
        \end{tabular*}
    \end{minipage}
    \vspace{-3pt}
\end{table}

\subsection{Long-Horizon Analysis}
\label{sec:analysis}

\paragraph{Settings.} We conduct long-horizon analysis on the RxR Val-Unseen split, which features complex instructions and long-horizon navigation episodes. Task complexity is quantified by the number of clauses in each instruction, delimited by ``\texttt{,}'' and ``\texttt{.}'', and categorized as: simple (1-3), medium (4-6), complex (7-10), and very complex (>10). We evaluate three model variants: the \ac{vlm} baseline, baseline with memory module, and the full \model. Beyond task complexity, we further categorize episodes by whether the instruction involves: (i) backtracking, (ii) explicit stopping conditions and long hallways, and (iii) passing landmarks. We report the success rate (SR) and navigation error (NE) metrics.

\begin{figure}[t!]
    \centering
    \includegraphics[width=\linewidth,pagebox=cropbox]{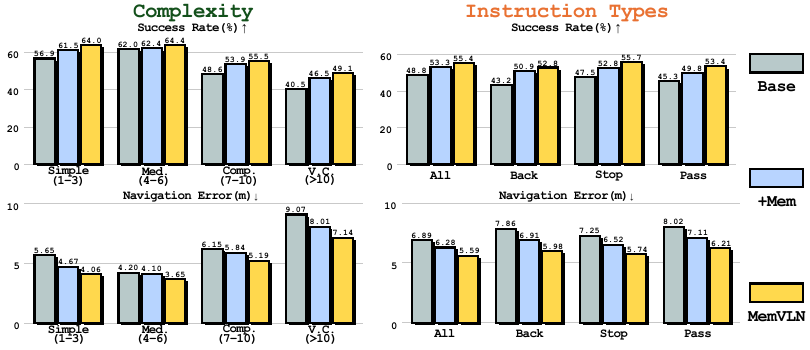}
    \caption{\textbf{Long-horizon analysis on RxR-CE Val-Unseen.} Success rate (top) and navigation error (bottom) are grouped by instruction complexity (left) and instruction type (right). Base denotes the \ac{vlm} baseline, and +Mem adds the memory module. Back, Stop, and Pass denote instructions involving backtracking, stop conditions and long hallways, and passing landmarks, respectively.}
    \label{fig:vln_analysis}
    \vspace{-3pt}
\end{figure}

\paragraph{Results.} In \cref{fig:vln_analysis}, memory raises SR and lowers NE across all complexity groups and instruction types; the full \model further improves both metrics. Across complexity levels, memory yields its largest SR gain on very complex instructions: SR rises from 40.5\% to 46.5\%, while NE falls from 9.07 to 8.01\,m. The full model reaches 49.1\% SR and 7.14\,m NE (+8.6\,pp SR over the baseline).

On backtracking episodes, memory raises SR from 43.2\% to 50.9\% and lowers NE from 7.86 to 6.91\,m; the full model reaches 52.8\% SR and 5.98\,m NE. Memory also gains 5.2\,pp and 4.5\,pp SR on stop-condition and landmark-passing subsets, respectively. These gains on complex instructions and backtracking highlight the value of episodic context for following extended instructions and revisiting observed locations.

\subsection{Real-world Experiment}
\label{sec:real_world}

\paragraph{Settings.} We deploy on G1 in one physical scene and collect 30 instruction-conditioned episodes with human corrections. We split these episodes into 24 post-training demonstrations and six held-out test episodes. We evaluate the \ac{vlm} baseline (Base), its memory-augmented variant (+Mem), the full \model, and the post-trained \model-PT, executing each held-out instruction five times for 30 trials per model. Post-training uses step-weighted behavior cloning, assigning larger weights to corrected steps. We report success rate (SR) and collision rate (CR), where CR is the fraction of trials involving physical contact, contact-induced blockage, or contact-induced termination.

\noindent
\begin{minipage}[c]{0.49\linewidth}
\textbf{Results.} In \cref{fig:real_world}, \model achieves 40.0\% SR (12/30 successful trials), up from Base's 16.7\% (5/30), while reducing CR from 56.7\% to 46.7\%. Memory alone raises SR to 30.0\% (four additional successes over Base). Compared with +Mem, the full model gains 10.0\,pp SR and lowers CR by 20.0\,pp. Post-training with the 24 demonstrations further raises SR from 40.0\% to 46.7\% (14/30 successful trials), while CR remains at 46.7\%.
\end{minipage}\hfill
\begin{minipage}[c]{0.45\linewidth}
    \centering
    \includegraphics[width=\linewidth]{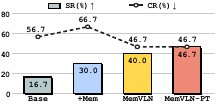}
    \captionof{figure}{\textbf{Real-world results on G1.}}
    \label{fig:real_world}
\end{minipage}

\paragraph{Inference Cost.} Fixed-input replay on one RTX~4090 yields mean policy latencies of 503.2\,ms for Base and 691.0\,ms for full \model. The memory module contributes 300 tokens to the \ac{llm} input, independent of route length. Profiling, deployment details, and per-instruction outcomes are provided in \cref{sec:supp_inference_cost,sec:supp_real_world_results}.

\subsection{Qualitative Results}
\label{sec:qualitative}

\begin{figure}[t!]
    \centering
    \includegraphics[width=\linewidth,pagebox=cropbox]{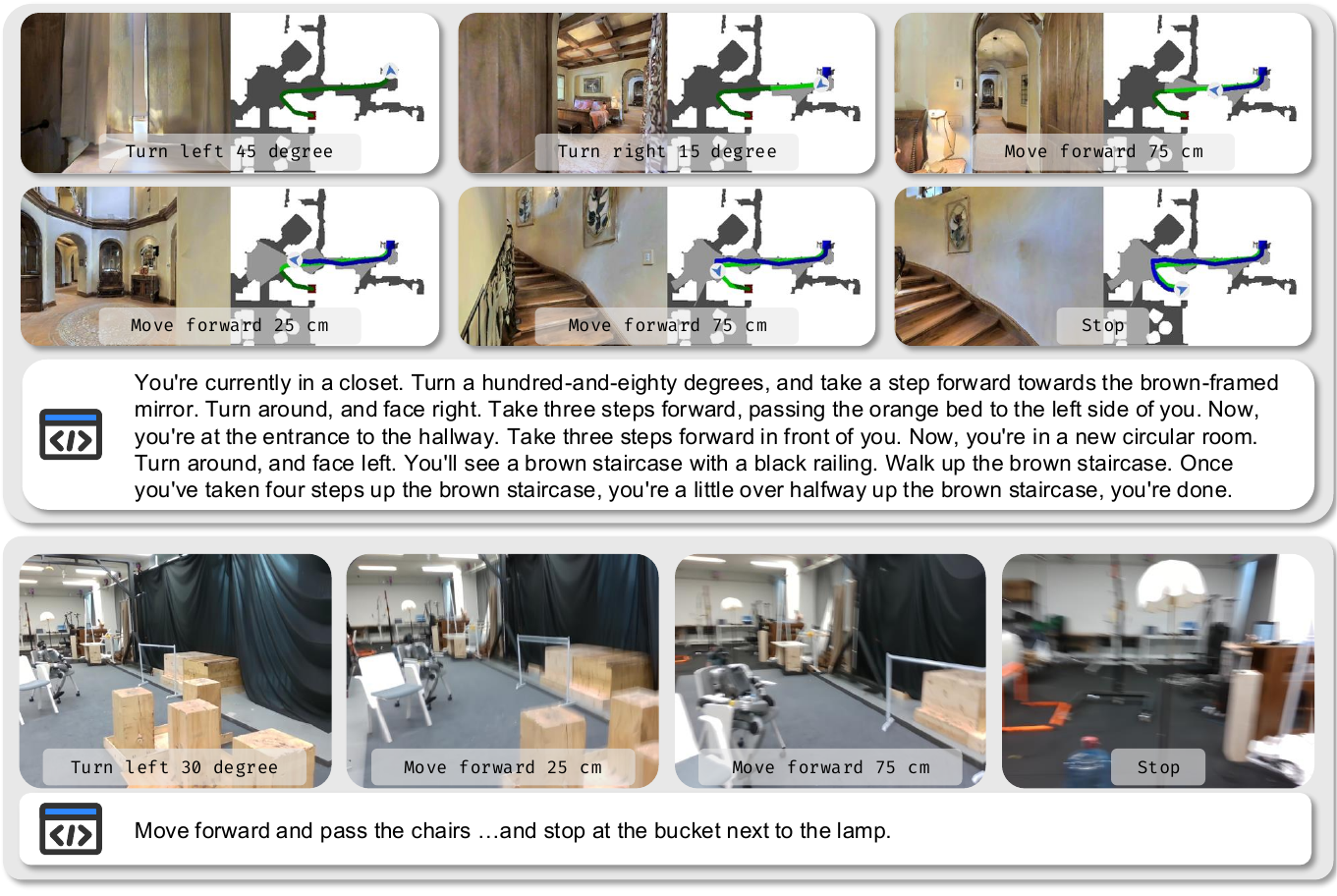}
    \caption{\textbf{Qualitative results in simulation and the real world.} Selected observations, actions, and instructions for RxR-CE (top, with top-down trajectories) and G1 (bottom).}
    \label{fig:vln_vis}
\end{figure}

\Cref{fig:vln_vis} shows an RxR-CE route from a closet through a hallway to a staircase, and a G1 rollout past chairs, stopping at the bucket beside a lamp. More examples are in \cref{sec:supp_qualit_res}.

\section{Conclusion}
\label{sec:conclusion}
\model combines plug-and-play episodic memory with a frozen pretrained geometry encoder for streaming \acl{vln}. Multi-granularity trajectory summarization trains the bounded memory representation, masking local visual tokens to reconstruct routes exclusively from episodic memory. With navigation training restricted to official R2R/RxR trajectories, \model achieves \sota performance on most reported VLN-CE metrics. Simulation analyses characterize memory's utility across instruction complexities and navigation scenarios. In one physical scene, post-training on 24 demonstrations improves success on six held-out episodes.


\section{Limitations and Future Work}
\label{sec:limitations}
First, inference-time retrieval and similarity-based consolidation are instruction-agnostic, so the bounded bank may discard task-relevant observations in long episodes. Second, while trajectory summarization supervises memory learning, broader training strategies remain under-explored~\citep{zheng2025flare}. Finally, the real-world study evaluates six held-out episodes in one scene, with substantial collision rates; cross-scene generalization is untested. Future work will investigate task-aware consolidation, broader deployment settings, and safety-aware post-training.


\clearpage
\subsection*{AI Use Statement}
\label{sec:ai_use_statement}

During manuscript revision, we used generative AI tools to assist
with literature discovery, drafting and editing selected passages
based on author-provided materials, discussing experimental design
and interpreting results, and developing and debugging research
code. These tools also assisted with figure preparation and LaTeX
formatting. The authors reviewed and revised the AI-assisted
material and take responsibility for the final manuscript and
reported results.

\bibliographystyle{iclr2027_conference}
\bibliography{reference}  

\clearpage
\appendix


\section{Additional Implementation Details}
\label{sec:supp_implementation}

This section complements the implementation details in the main paper with the training protocol and memory configuration used for the default \model configuration.

\paragraph{Data Statistics.}
We summarize the training data statistics in \cref{tab:supp_data_statistics}. Each instruction-conditioned demonstration is treated as one episode. For simulation training, the number of navigation samples counts the step-wise action-prediction targets after action merging; the trajectory summarization data is derived from the same episodes. For real-world post-training, we collect 24 demonstrations using an Intel RealSense D455 RGB-D camera mounted on the G1 robot for training. The D455 RGB sensor has a nominal horizontal field of view of $90^\circ$. Its RGB frames are captured at $1280 \times 720$ and resized to $640 \times 480$ before model processing to match the simulation input resolution.


\begin{table}[h!]
\centering
\caption{\textbf{Training data statistics.} Navigation samples count step-wise action-prediction targets after action merging. Simulation sample counts are rounded to the nearest thousand.}
\label{tab:supp_data_statistics}
\resizebox{\textwidth}{!}{
\begin{tabular}{lcrrccc}
\toprule
\multirow{2}{*}[-0.3em]{Setting} &
\multirow{2}{*}[-0.3em]{Dataset} &
\multicolumn{2}{c}{Training Data} &
\multicolumn{3}{c}{RGB Input} \\
\cmidrule(lr){3-4}
\cmidrule(lr){5-7}
& & Episodes & Nav. Samples & Resolution & HFOV & Camera \\
\midrule
\multirow{2}{*}{Simulation} & R2R Train & 10{,}819 & $\sim$334K & $640 \times 480$ & $90^\circ$ & Habitat-Sim \\
& RxR Train & 19{,}996 & $\sim$1.02M & $640 \times 480$ & $90^\circ$ & Habitat-Sim \\
\midrule
Real-world PT & G1 demonstrations & 24 & 1{,}053 & $640 \times 480$ & $90^\circ$ & Intel RealSense D455 \\
\bottomrule
\end{tabular}}
\end{table}

\paragraph{Streaming Episode Sampler.}
Training a stateful memory policy requires temporally coherent data loading. Standard random sampling breaks an episode into unrelated steps, preventing the memory bank from being updated in the same chronological order encountered at inference time. We therefore employ a streaming episode sampler that assigns complete episodes to each device and feeds the steps within each episode sequentially. This protocol allows each device to maintain continuous memory streams while supporting distributed training and task mixing.

\paragraph{Default Memory Configuration.}
Following the memory module formulation in the main paper, the memory bank stores the visual feature $f_t$ of each observation together with its timestep $t$ and relative pose $z_t$; at policy inference, $z_t$ is derived from the executed action history. We denote the maximum number of retained frame entries as $N_b$. Each frame is spatially compressed to at most $N_f$ patch tokens using ToMe~\citep{bolya2023token}. A new observation is first appended to the bank. If this increases the bank size to $N_b+1$, frame merging consolidates the two consecutive entries with the highest average patch-level cosine similarity, reducing the bank size by one and restoring the capacity constraint. The memory perceiver then uses $N_m$ learnable queries to retrieve a fixed-size set of memory tokens $m_t$. Before each memory readout, the current visual feature is added to the bank; the queries are conditioned only on the current timestep and action-derived relative pose, while the instruction is introduced downstream by the \ac{llm}. Since each query produces one output token, $N_m$ is also the number of memory tokens injected into the \ac{vlm}. The default configuration uses $N_f=100$, $N_b=100$, and $N_m=300$. Additional settings are summarized in \cref{tab:supp_memory_hyperparameters}.

\paragraph{Simulation Training.}
As described in the main paper, we adopt Qwen3-VL-4B~\citep{bai2025qwen3} as the \ac{vlm} backbone and StreamVGGT-1B~\citep{zhuo2026streaming} as the geometry encoder. The visual encoder and geometry encoder remain frozen. We adapt the \ac{llm} with LoRA~\citep{hu2022lora}, while training the memory module and geometry adapter from scratch. More details are summarized in \cref{tab:supp_training_hyperparameters}.

\begin{table}[htbp]
\centering
\begin{minipage}[t]{0.49\linewidth}
\vspace{0pt}
\centering
\caption{\textbf{Default memory module configuration.}}
\label{tab:supp_memory_hyperparameters}
\footnotesize
\setlength{\tabcolsep}{4pt}
\renewcommand{\arraystretch}{1.08}
\begin{tabularx}{\linewidth}{@{}>{\raggedright\arraybackslash}X>{\raggedleft\arraybackslash}p{0.44\linewidth}@{}}
\toprule
\textbf{Hyperparameter} & \textbf{Setting} \\
\midrule
Local observation window $k$ & 16 \\
Max tokens per frame $N_f$ & 100 \\
Memory bank capacity $N_b$ & 100 \\
Memory token count $N_m$ & 300 \\
\midrule
Frame consolidation & Sim-based merging \\
Spatial embedding & Fourier features \\
Temporal embedding & Sinusoidal embedding \\
\midrule
ToMe layers & 2 \\
ToMe attention heads & 16 \\
Perceiver layers & 2 \\
Perceiver attention heads & 16 \\
\bottomrule
\end{tabularx}
\end{minipage}
\hfill
\begin{minipage}[t]{0.49\linewidth}
\vspace{0pt}
\centering
\caption{\textbf{\model training hyperparameters.}}
\label{tab:supp_training_hyperparameters}
\footnotesize
\setlength{\tabcolsep}{4pt}
\renewcommand{\arraystretch}{1.08}
\begin{tabularx}{\linewidth}{@{}>{\raggedright\arraybackslash}X>{\raggedleft\arraybackslash}p{0.44\linewidth}@{}}
\toprule
\textbf{Category} & \textbf{Setting} \\
\midrule
Hardware & 8 $\times$ A100 80G GPUs \\
Precision & BF16 \\
Training epochs & 1 \\
Per-device batch size & 1 \\
Gradient accumulation steps & 1 \\
Effective batch size & $1 \times 8 \times 1 = 8$ \\
\midrule
Optimizer & AdamW \\
Base learning rate & $2 \times 10^{-5}$ \\
Weight decay & 0.05 \\
Warmup steps & 400 \\
\midrule
LoRA rank / alpha & 128 / 256 \\
\bottomrule
\end{tabularx}
\end{minipage}
\end{table}

\paragraph{Real-world Post-training.}
The real-world demonstrations record both the model prediction and the human-corrected executed action. We post-train the policy using the same token-level cross-entropy objective as in simulation, while assigning a larger sample weight to steps requiring correction:
\begin{equation}
    \mathcal{L}_{\text{PT}}
    =
    \frac{\sum_t w_t \mathcal{L}_t}{\sum_t w_t},
    \qquad
    \mathcal{L}_t
    =
    -\log p(a_t^* \mid m_t,\mathcal{H}_t,\mathcal{T}).
\end{equation}
We set $w_t$ to 3.0 for corrected non-stop actions, 0.5 for actions accepted without correction, and 2.0 for stop actions. This correction-weighted behavior cloning objective focuses adaptation on states where the deployed policy requires human intervention.

\section{Inference Latency and Computational Cost}
\label{sec:supp_inference_cost}

\paragraph{Evaluation Protocol.}
We measure system-level cost using fixed-input policy replay on one NVIDIA RTX~4090 with batch size 1 and the configured BF16/FlashAttention~2 inference setup. We compare the Base policy with the complete \model pipeline to measure the combined system overhead of StreamVGGT and episodic memory. The replay uses three frozen R2R episode traces comprising 102 fixed policy decisions. Each variant is evaluated in two fresh processes after the same 20-decision warm-up and state reset, yielding 204 timed observations per variant, including 168 observations in which the recent-context window contains 16 frames. Timing starts from RGB arrays in CPU memory and includes message construction, model preprocessing, host-to-device transfer, policy generation, decoding, and action parsing. It excludes sensor capture, Wi-Fi transfer, environment stepping, and the low-level robot controller. This fixed-input diagnostic characterizes policy cost and is not used to compute navigation metrics.

\begin{table}[h!]
    \centering
    \caption{\textbf{Fixed-replay inference latency and peak resource usage on one RTX~4090.} Latency statistics pool two fresh-process repeats of the same 102 fixed decisions; $N$ denotes timed observations rather than independent episodes. The 16-frame recent-context subset contains 168 timed observations per variant. Policy rate is $1000/\text{mean latency (ms)}$. Resource rows report the maximum across valid repeats. Process GPU is the peak PID-scoped memory reported by \texttt{nvidia-smi}, and RSS is the peak host resident set size of the evaluation process.}
    \label{tab:supp_inference_cost}
    \begin{tabular}{lcc}
        \toprule
        Metric & Base & Full \model \\
        \midrule
        Timed observations, $N$ & 204 & 204 \\
        Mean latency (ms)$\downarrow$ & 503.2 & 691.0 \\
        P50 latency (ms)$\downarrow$ & 535.2 & 704.3 \\
        P95 latency (ms)$\downarrow$ & 550.0 & 779.8 \\
        Policy rate (Hz)$\uparrow$ & 1.99 & 1.45 \\
        \midrule
        16-frame observations, $N$ & 168 & 168 \\
        16-frame mean latency (ms)$\downarrow$ & 534.9 & 731.7 \\
        \midrule
        CUDA peak allocated (GiB)$\downarrow$ & 10.725 & 17.104 \\
        CUDA peak reserved (GiB)$\downarrow$ & 11.234 & 21.078 \\
        Process GPU peak (GiB)$\downarrow$ & 11.69 & 21.54 \\
        Host RSS peak (GiB)$\downarrow$ & 3.401 & 8.666 \\
        \bottomrule
    \end{tabular}
\end{table}

\paragraph{End-to-End Latency.}
As shown in \cref{tab:supp_inference_cost}, the complete pipeline increases mean policy latency from 503.2 to 691.0~ms (37.3\%), corresponding to a sequential high-level policy rate of 1.45~Hz on the tested GPU. For the 16-frame recent-context subset, mean latency is 534.9~ms for Base and 731.7~ms for Full \model, a 36.8\% increase.

\paragraph{Fixed Token Interface.}
For one R2R decision with a 16-frame recent-context window, Base receives 2{,}819 input tokens and Full \model receives 3{,}126. In both cases, 15 history frames are padded to 16 and grouped into eight temporal units; these units and the current image each contribute 300 visual tokens, for $(8+1)\times300=2{,}700$ visual tokens. Full \model additionally contributes 300 fixed memory tokens and 7 associated prompt/special tokens, a fixed increase of 307 tokens under the evaluated prompt format. Other text tokens depend on the instruction. The bounded raw bank is not concatenated into the \ac{vlm} context, so this memory-side input increment does not grow with route length.

\paragraph{Control Frequency.}
The measured mean latency in \cref{tab:supp_inference_cost} corresponds to a sequential high-level policy rate of 1.45~Hz under the tested configuration. The real-robot client uses a separate low-level PID control thread to send commands through the Unitree SDK, nominally configured at approximately 30~Hz. The RTX~4090 benchmark measures high-level policy inference. We use \emph{streaming} to denote causal online processing with bounded state.

\paragraph{Training Configuration.}
The main paper reports that the simulation model has approximately 1B trainable parameters and is trained for one epoch, taking approximately two days on 8$\times$A100 80G GPUs. Additional training hyperparameters are summarized in \cref{tab:supp_training_hyperparameters}.

\section{Additional Results}
\label{sec:supp_additional_results}

\subsection{Memory Capacity}
\label{sec:supp_memory_capacity}

\Cref{tab:supp_memory_capacity} reports the memory bank capacity and memory-token count ablations on both benchmarks. The default configuration uses $N_b=100$, $N_m=300$, and $N_f=100$ patch tokens per stored frame. The zero-capacity setting retains the geometry encoder and trajectory-summary supervision.

\begin{table}[t!]
    \centering
    \caption{\textbf{Memory capacity ablations on R2R-CE and RxR-CE.}}
    \label{tab:supp_memory_capacity}
    \small
    \setlength{\tabcolsep}{4pt}
    \textbf{(a) Memory bank capacity}\par
    \begin{tabular*}{\linewidth}{@{\extracolsep{\fill}}*{9}{c}@{}}
        \toprule
        \multirow{2}{*}[-0.3em]{$N_b$} & \multicolumn{4}{c}{R2R Val-Unseen} & \multicolumn{4}{c}{RxR Val-Unseen} \\
        \cmidrule(lr){2-5} \cmidrule(lr){6-9}
        & NE$\downarrow$ & OSR$\uparrow$ & SR$\uparrow$ & SPL$\uparrow$ & NE$\downarrow$ & SR$\uparrow$ & SPL$\uparrow$ & nDTW$\uparrow$ \\
        \midrule
        0 & 5.87 & 57.4 & 50.9 & 47.2 & 5.72 & 54.9 & \textbf{49.0} & 64.5 \\
        100 & \textbf{5.19} & \textbf{61.2} & \textbf{55.0} & \textbf{50.5} & \textbf{5.59} & \textbf{55.4} & 48.4 & \textbf{64.6} \\
        200 & 6.24 & 58.7 & 50.2 & 45.7 & 6.15 & 54.3 & 48.0 & 61.5 \\
        \bottomrule
    \end{tabular*}
    \par\vspace{0.65em}
    \textbf{(b) Memory-token count}\par
    \begin{tabular*}{\linewidth}{@{\extracolsep{\fill}}*{9}{c}@{}}
        \toprule
        \multirow{2}{*}[-0.3em]{$N_m$} & \multicolumn{4}{c}{R2R Val-Unseen} & \multicolumn{4}{c}{RxR Val-Unseen} \\
        \cmidrule(lr){2-5} \cmidrule(lr){6-9}
        & NE$\downarrow$ & OSR$\uparrow$ & SR$\uparrow$ & SPL$\uparrow$ & NE$\downarrow$ & SR$\uparrow$ & SPL$\uparrow$ & nDTW$\uparrow$ \\
        \midrule
        0 & 5.87 & 57.4 & 50.9 & 47.2 & 5.72 & 54.9 & \textbf{49.0} & 64.5 \\
        300 & \textbf{5.19} & \textbf{61.2} & \textbf{55.0} & \textbf{50.5} & \textbf{5.59} & \textbf{55.4} & 48.4 & \textbf{64.6} \\
        600 & 6.75 & 53.9 & 45.1 & 40.8 & 6.64 & 50.4 & 44.0 & 59.3 \\
        \bottomrule
    \end{tabular*}
\end{table}

\subsection{Long-Horizon Analysis Protocol}
\label{sec:supp_analysis_protocol}

We apply the same instruction-based grouping to all three model variants over 3,669 RxR-CE Val-Unseen episodes in \cref{fig:vln_analysis}. For instruction complexity, we count the total occurrences of commas and periods in each instruction as a punctuation-based proxy. The four bins contain 1--3, 4--6, 7--10, and more than 10 punctuation marks, comprising 239, 819, 1,113, and 1,498 episodes, respectively.

Instruction-type subsets are selected by case-insensitive keyword matching with word boundaries. The Back subset matches \texttt{turn around}, \texttt{back}, or \texttt{behind}, yielding 1,272 episodes. The Stop subset matches \texttt{stop}, \texttt{wait}, \texttt{end}, \texttt{hallway}, \texttt{hall}, or \texttt{corridor}, yielding 2,072 episodes. The Pass subset matches \texttt{pass}, \texttt{past}, \texttt{through}, or \texttt{cross}, yielding 1,410 episodes. An instruction enters a subset when it contains at least one corresponding keyword or phrase. These subsets are constructed independently and can overlap; All denotes the complete set of 3,669 episodes.

\subsection{Real-world Results}
\label{sec:supp_real_world_results}

\paragraph{Evaluation Protocol.}
We provide a detailed breakdown of the real-world G1 evaluation summarized in \cref{sec:real_world}. All 30 instruction-conditioned episodes were collected in the same physical scene and split at the episode level: 24 demonstrations are used for post-training and six distinct episodes are held out for evaluation. We evaluate Base, +Mem, MemVLN, and MemVLN-PT, repeating each held-out instruction five times for 30 trials per method. In addition to success rate (SR), we report collision rate (CR), the fraction of trials involving physical contact, contact-induced blockage, or contact-induced termination; CR measures contact safety independently of task completion.

\paragraph{Deployment Setup.}
During real-world deployment, policy inference runs on a host workstation equipped with a single NVIDIA RTX 4090 GPU. RGB observations captured by the robot-mounted Intel RealSense D455 are transmitted to the host over Wi-Fi; after the host completes model inference, the predicted action is sent back to the G1 robot for execution.

\begin{table}[t!]
\centering
\captionof{table}{\textbf{Real-world evaluation instructions.} We use the instruction IDs in subsequent results.}
\label{tab:supp_realworld_instructions}
\footnotesize
\begin{tabularx}{\textwidth}{@{}l>{\raggedright\arraybackslash}X@{}}
    \toprule
    ID & Instruction \\
    \midrule
    I1 & Now you are facing a brown curtain. Turn right and walk pass an armchair towards the fire extinguisher. Turn left and walk toward the glass door in front of you to the right, that's your destination. \\
    I2 & Now you are facing the reception. On your left you should see some brown sofas. Walk past the sofas to the handwriting on the wall. Stop at the door on your right. \\
    I3 & Now you are in a corridor. Keep moving forward. Once you pass the bookshelf on the right side, turn right and go to the sink in the corner. \\
    I4 & Go straight towards the glass windows and past the sofas on your right. After you reach the windows, turn right. Keep moving forward and stop at the second door on your right. \\
    I5 & Now you are in a corridor. On your left, there are some white cabinets. Go past them and then turn left. You should see a red fire extinguisher on the ground. That box is your destination. \\
    I6 & Now you are in a lab. On your right there are some wooden stakes. Move forward and pass the chairs in front of you. You should see a robot sitting on a chair. Walk past the robot and stop at the bucket next to the lamp. \\
    \bottomrule
\end{tabularx}
\end{table}

\paragraph{Overall Performance.}
As shown in \cref{fig:real_world}, Base succeeds in 5 of 30 runs, corresponding to 16.7\% SR. Adding the memory module increases success to 9 of 30 runs (30.0\% SR), and the full MemVLN model further improves success to 12 of 30 runs (40.0\% SR). After real-world post-training with 24 demonstrations, MemVLN-PT succeeds in 14 of 30 runs (46.7\% SR), giving two additional successful runs over MemVLN. CR remains unchanged at 46.7\% for MemVLN and MemVLN-PT.

\begin{table}[t!]
\centering
\captionof{table}{\textbf{Per-instruction real-world results.} Each cell reports SR / CR (\%) over five repeated trials for the corresponding instruction and method. Bold marks the best nonzero SR and lowest CR in each row.}
\label{tab:supp_realworld_per_instruction}
\small
\begin{tabular}{lcccc}
    \toprule
    ID & Base & +Mem & MemVLN & MemVLN-PT \\
    \midrule
    I1 & 20 / 40 & 40 / 60 & \textbf{80} / \textbf{20} & 60 / 80 \\
    I2 & 0 / 60 & 0 / 60 & \textbf{20} / 60 & \textbf{20} / \textbf{40} \\
    I3 & 40 / 60 & \textbf{80} / 60 & 40 / \textbf{40} & \textbf{80} / 60 \\
    I4 & 0 / 100 & 0 / 100 & 0 / \textbf{60} & 0 / \textbf{60} \\
    I5 & 20 / \textbf{20} & 0 / 40 & 20 / 40 & \textbf{60} / 40 \\
    I6 & 20 / 60 & 60 / 80 & \textbf{80} / 60 & 60 / \textbf{0} \\
    \bottomrule
\end{tabular}
\end{table}

\paragraph{Instruction-level Analysis.}
As shown in \cref{tab:supp_realworld_per_instruction}, MemVLN-PT improves SR over MemVLN on I3 and I5, leaves it unchanged on I2 and I4, and decreases it on I1 and I6. Post-training lowers CR on I2 and I6 and increases it on I1 and I3. All four methods obtain 0\% SR on I4, which requires selecting the second door on the right after passing an intermediate door.

\paragraph{Remaining Challenges.}
Collision remains a challenge: MemVLN and MemVLN-PT both have a CR of 46.7\%. Post-training improves aggregate SR but leaves aggregate CR unchanged. Future work will explore contact-aware control and richer supervision for fine-grained spatial grounding.

\subsection{More Qualitative Results}
\label{sec:supp_qualit_res}

\begin{figure}[htbp]
    \centering
        \includegraphics[width=\linewidth]{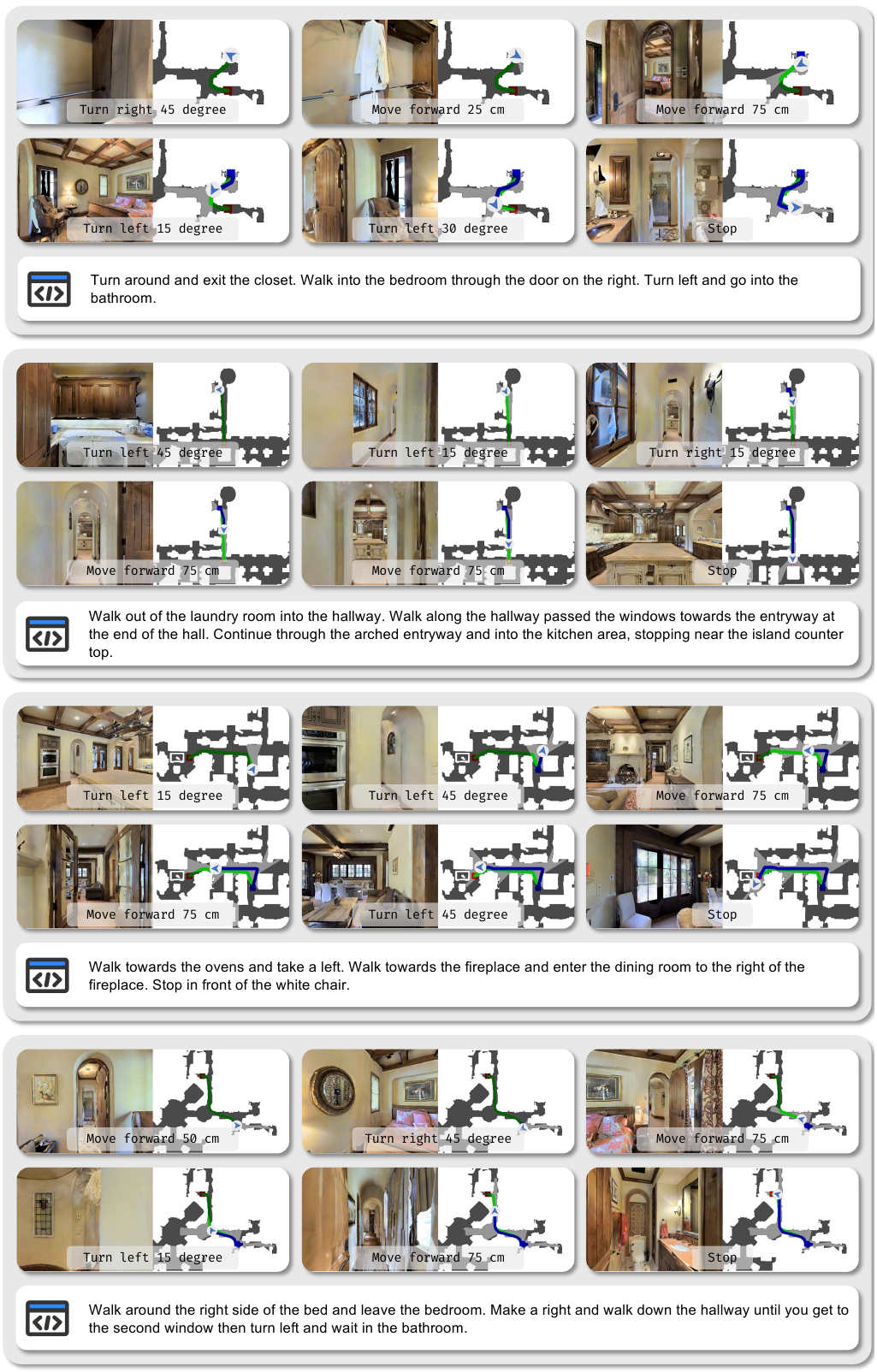}
        \caption{
        \textbf{Qualitative results of \model on R2R-CE.}
        }
    \label{fig:supp_r2r}
\end{figure}

\begin{figure}[p]
    \centering
        \includegraphics[width=\linewidth,height=0.90\textheight,keepaspectratio]{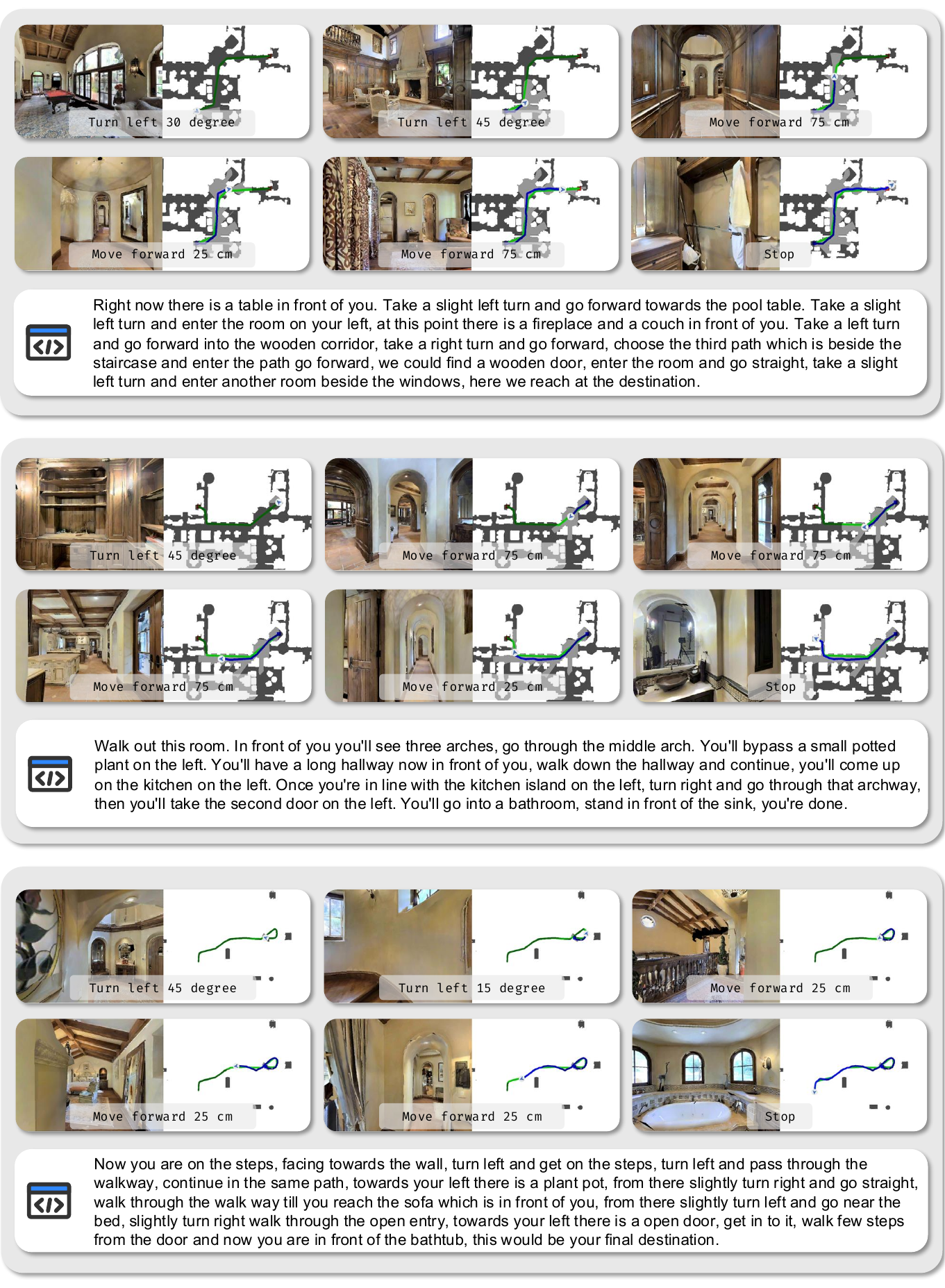}
        \caption{
        \textbf{Qualitative results of \model on RxR-CE.}
        }
    \label{fig:supp_rxr}
\end{figure}

\paragraph{Simulation Results.}
\Cref{fig:supp_r2r,fig:supp_rxr} show R2R-CE and RxR-CE navigation rollouts, including selected observations and actions along multi-step routes.

\begin{figure}[htbp]
    \centering
    \includegraphics[width=\linewidth]{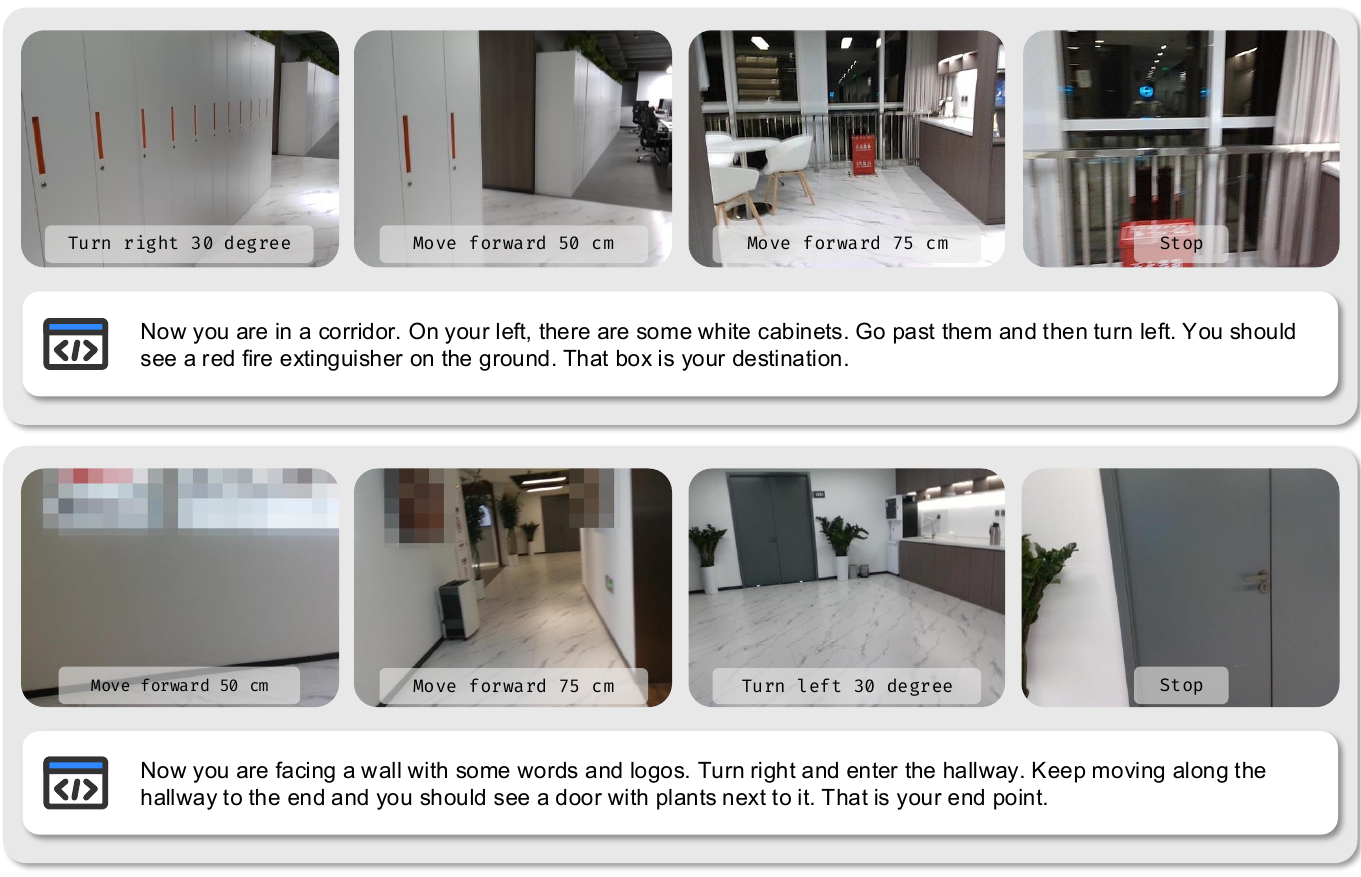}
    \caption{\textbf{Qualitative results of \model in the real world.} Selected observations and actions from two G1 navigation rollouts, together with their navigation instructions.}
    \label{fig:supp_real}
\end{figure}

\clearpage
\paragraph{Real-world Results.}
\Cref{fig:supp_real} shows two additional G1 navigation rollouts with their instructions, selected observations, and actions.

\end{document}